\documentclass[smallabstract,smallcaptions]{dccpaper}

\usepackage{epsfig}
\usepackage{cite}
\usepackage{amsmath}
\usepackage{amssymb}
\usepackage{color}
\usepackage{url}
\usepackage{times}
\usepackage{epsfig}

\usepackage{algorithm}
\usepackage{algorithmic}
\usepackage{balance}
\usepackage{bm}
\usepackage{caption}
\usepackage{subcaption}

\PassOptionsToPackage{hyphens,spaces,obeyspaces}{url}\usepackage{hyperref}

\newlength{\figurewidth}
\newlength{\smallfigurewidth}

\begin{document}

\title
{\large
\textbf{DRASIC: Distributed Recurrent Autoencoder for Scalable Image Compression}
\thanks{This work was supported in part by Office of Naval Research Grant No. N00014-18-1-2244. We provide our implementation at \url{https://github.com/dem123456789/Distributed-Recurrent-Autoencoder-for-Scalable-Image-Compression}}
}

\author{%
Enmao Diao$^{\ast}$, Jie Ding$^{\dag}$, and Vahid Tarokh$^{\ast}$\\[0.5em]
{\small\begin{minipage}{\linewidth}
\begin{center}
\begin{tabular}{cc}
$^{\ast}$Duke University &
$^{\dag}$University of Minnesota-Twin Cities\\
Durham, NC, 27701, USA &
Minneapolis, MN 55455, USA\\
\url{enmao.diao@duke.edu}&\url{dingj@umn.edu}\\
\url{vahid.tarokh@duke.edu}&
\end{tabular}
\end{center}\end{minipage}}
}

\maketitle
\thispagestyle{empty}

\begin{abstract}
We propose a new architecture for distributed image compression from a group of distributed data sources. The work is motivated by practical needs of data-driven codec design, low power consumption, robustness, and data privacy. The proposed architecture, which we refer to as Distributed Recurrent Autoencoder for Scalable Image Compression (DRASIC), is able to train distributed encoders and one joint decoder on correlated data sources. Its compression capability is much better than the method of training codecs separately. Meanwhile, the performance of our distributed system with 10 distributed sources is only within 2 dB peak signal-to-noise ratio (PSNR) of the performance of a single codec trained with all data sources. We experiment distributed sources with different correlations and show how our data-driven methodology well matches the Slepian-Wolf Theorem in Distributed Source Coding (DSC). To the best of our knowledge, this is the first data-driven DSC framework for general distributed code design with deep learning.
\end{abstract}

%%%%%%%%% BODY TEXT
\section{Introduction}
It has been shown by a variety of previous works that deep neural networks (DNN) can achieve comparable results as classical image compression techniques \cite{toderici2015variable,balle2016end,gregor2016towards,toderici2017full,theis2017lossy,johnston2017improved,liu2018cnn,li2018learning,mentzer2018conditional}. Most of these methods are based on autoencoder networks and quantization of bottleneck representations. These models usually rely on entropy codec to further compress codes. Moreover, to achieve different compression rates it is unavoidable to train multiple models with different regularization parameters separately, which is often computationally intensive.

In this work, we are motivated to develop an architecture that has the following advantages. 
First, unlike classical distributed source coding (DSC) which requires customized code design for different scenarios~\cite{xiong2004distributed}, a data-driven distributed compression framework can handle nontrivial distribution of image sources with arbitrary correlations. Second, the computation complexity of encoders (e.g. mobile devices) can be transferred to the decoder (e.g. a remote server). Such a system of low complexity encoders can be used in a variety of application domains, such as multi-view video coding \cite{girod2005distributed}, sensor networks \cite{xiong2004distributed}, and under-water image processing where communication bandwidth and computational power are quite restricted \cite{stojanovic2009underwater,schettini2010underwater}. Third, the distributed framework can be more robust against heterogeneous noises or malfunctions of encoders, and such robustness can be crucial in, e.g., unreliable sensor networks \cite{girod2005distributed,ishwar2005rate,xiao2006distributed}.
Last but not least, the architecture is naturally scalable in the sense that codes can be decoded at more than one compression quality level, and it allows efficient coding of correlated sources which are not physically co-located. This is especially attractive in video streaming applications \cite{guillemot2007distributed,gehrig2008distributed}.

It is tempting to think that splitting raw data for different encoders compromises the compression quality. It is thus natural to ask this question: Can distributed encoders perform as well as a single encoder trained with all data sources together?
A positive answer from a theoretical perspective was given in the context of information theory, where DSC is an important problem regarding the compression of multiple correlated data sources. The Slepian-Wolf Theorem shows that lossless coding of two or more correlated data sources with separate encoders and a joint decoder can compress data as efficiently as the optimal coding using a joint encoder and decoder \cite{slepian1973noiseless,cover1975proof}. The extension to lossy compression with Gaussian data sources was proposed as Wyner-Ziv Theorem \cite{wyner1976rate}. Although these theorems were published in 1970s, it was after about 30 years that practical applications such as Distributed Source Coding Using Syndromes (DISCUS) emerged \cite{pradhan2003distributed}. One of the main advantages of DSC is that the computation complexity of the encoder is transferred to the decoder. A system architecture with low complexity encoders can be a significant advantage in applications such as  multi-view video coding and sensor networks \cite{girod2005distributed,xiong2004distributed}. 

\begin{figure*}
\begin{center}
\vspace{-0.2cm}
\includegraphics[width=0.5\linewidth]{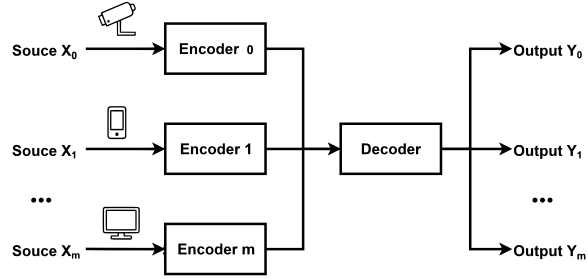}
\vspace{-0.2cm}
\end{center}
   \caption{Illustration of Deep Distributed Source Coding.}
\label{fig_0}
\end{figure*}

\begin{figure*}
\begin{center}
\vspace{-0.2cm}
\includegraphics[width=0.7\linewidth]{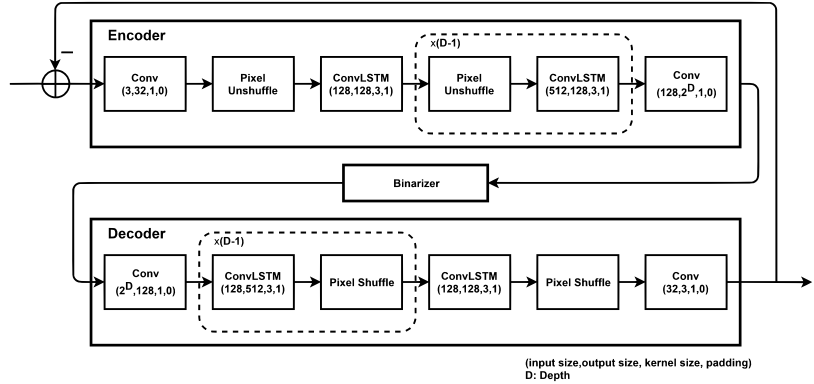}
\vspace{-0.2cm}
\end{center}
   \caption{Illustration of Recurrent Autoencoder for Scalable Image Compression.}
\label{fig_1}
\end{figure*}

Motivated by the theoretical development of DSC, in this work we propose a DNN architecture that consists of distributed encoders and a joint decoder (illustrated in Fig.~\ref{fig_0} and \ref{fig_1}). We show that distributed encoders can perform as well as a single encoder trained with all data sources together.
Our proposed DSC framework is data-driven by nature, and it can be applied to distributed data even with unknown correlation structure.

The paper is outlined below. We review previous related works in Section 2. We describe our proposed architecture for general image compression and its basic modules in Subsections 3.1-3.4. Then we elaborate the Deep Distributed Source Coding framework in Subsection 3.5. Experimental results are shown in Section 4, followed by conclusions in Section 5.
%------------------------------------------------------------------------
\section{Related Work}
Though there has been a variety of research on lossy data compression in the past few decades, little attention has been paid to a systematic approach for general and practical distributed code design, especially in the presence of an arbitrary number of nontrivial data sources with arbitrary correlations \cite{xiong2004distributed}. 
A main motivation of this work is to attempt to replace the practical hand-crafted code design with data-driven approaches. 
To our best knowledge, what we propose is the first data-driven DSC architecture. 
Unlike hand-crafted quantizers, our neural network-based quantizers show that the correlations among different data sources can be exploited by the model parameters. Inspired by DSC, We empirically show that it is possible to approach the theoretical limit with our methodology. 

\subsection{Image compression with Deep Learning}
There exist a variety of classical codecs for lossy image compression. Although the JPEG standard \cite{wallace1992jpeg} was developed thirty years ago, it is still the most widely used image compression method. Several extensions to JPEG including JPEG2000 \cite{skodras2001jpeg}, WebP \cite{google2010webp} and BPG \cite{bellard2014bpg} have been developed. Most of these classical codecs rely on a quantization matrix applied to the coefficients of discrete cosine transform or wavelet transform.

Common deep neural network architecture for image compression are auto-encoders including non-recurrent autoencoders \cite{balle2016end,theis2017lossy,li2018learning,mentzer2018conditional} and recurrent autoencoders \cite{toderici2015variable,toderici2017full,johnston2017improved}. Non-recurrent autoencoders use entropy codec to encode quantized bottleneck representations, and recurrent models introduce incremental binarized codes at each compression quality. The generated codes of non-recurrent models is not scalable and their performance heavily relies on the conditional generative model like PixelCNN \cite{van2016conditional} which arithmetic coding can take advantage of \cite{li2018learning,mentzer2018conditional}. Recurrent autoencoders, on the other hand, can reconstruct images at lower compression qualities with the subset of high quality codes. Other notable variations include adversarial training \cite{rippel2017real}, multi-scale image compression \cite{nakanishi2018neural}, and generalized divisive normalization (GDN) layers \cite{balle2016end}. Another challenge is to well define the derivative of quantizations of bottleneck representations. \cite{balle2016end} replaced non-differentiable quantization step with a continuous relaxation by adding uniform noises. \cite{toderici2015variable}, on the other hand, used a stochastic form of binarization.

\subsection{Distributed Source Coding}
Our methodology is inspired by the information-theoretic results on DSC which have been established since 1970s. The Slepian-Wolf \cite{slepian1973noiseless} Theorem shows that two correlated data sources encoded separately and decoded jointly can perform as well as joint encoding and decoding, and outperform separate encoding and separate decoding. The striking result indicates that as long as the codes are jointly decoded, there can be no loss in coding efficiency even the codes are separately encoded. Cover \cite{cover1975proof} generalizes the achievability of Slepian-Wolf coding to arbitrary number of correlated sources. \cite{wyner1976rate} Coding gives a rate-distortion curve as an extension to lossy cases. Some researchers have also shown the applicability of DSC on still images \cite{dikici2005distributed}. In practical applications, low complexity video encoding benefits from the DSC framework which can transfer the complexity of encoder to decoder \cite{puri2002prism,aaron2002wyner}. Scalable Video Coding can also be incorporated with DSC \cite{xu2006layered}. These proposed methods indicate the feasibility of DSC in our problem setting.

\section{Methods}
In this section, we first describe the recurrent autoencoder for scalable image compression used in our work. We will then describe how this Deep Learning architecture is used in Distributed Source Coding framework.

\subsection{Network Architecture}
Our compression network consists of an encoder, a binarizer, and a decoder. The activation function following each Convolutional Neural Network (CNN) module is $\tanh$. For the first iteration of our model, the input images are initially encoded and transformed into $(-1,1)$ by $\tanh$ activation function. Binary codes are quantized from bottleneck representations. The decoder then reconstructs images based on the received binary codes. Finally, we compute the residual difference between the original input images and the reconstructed output images. At the next iteration, the residual difference is feedback as the new input for our model. This procedure is repeated multiple iterations to gain more codes for better reconstruction performance. Therefore, the reconstructed images at each iteration are the sum of output reconstructions from previous and current iterations. The dependencies among iterations are modeled by recurrent models like ConvLSTM. We iterate $16$ times to generate scalable codes. Compared to non-scalable codes which require new set of codes at each compression quality, scalable codes are able to reconstruct images at lower compression quality by using the subset of codes. This is especially attractive in video streaming applications \cite{guillemot2007distributed,gehrig2008distributed}.

Consider dataset $X = \{x\}^{N}$ consisting of $N$ i.i.d. samples of some continuous or discrete variables $x$. The data generating process is unknown. Autoencoders for compression and reconstruction can be formulated in the following way. Data can be compressed with a neural network-based encoder $f(x;\theta)$ into quantized codes $\tilde{z}$ and reconstructed with a decoder $g(\tilde{z};\phi)$. We can binarize bottleneck representations $z$ and control the compression quality by varying its channel sizes. The loss function $\mathcal{L}(x,\tilde{x})$ is minimized with respect to the model parameters $\theta$ and $\phi$.
\begin{align}
z = f(x;\theta),
\tilde{z} = \text{Binarize}&(z),
\tilde{x} = g(\tilde{z};\phi),\\
\text{Minimize } &\mathcal{L}(x,\tilde{x})
\end{align}
Deep recurrent autoencoder gradually increases compression quality by creating a correlated residual sequence from the difference between the input and output of our model. The advantage of recurrent model is that we can use a subset of generated codes to reconstruct images at lower compression qualities. Classical autoencoders, on the contrary, not only have to train multiple networks with different penalty coefficients for rate-distortion loss but also have to generate different codes for different compression quality. Suppose $T$ iterations are used, we can formulate the recurrent autoencoder in the following way.
\begin{align}
z_t &= f(x_t;\theta),
\tilde{z}_t = \text{Binarize}(z_t),\\
\tilde{x}_t &= g(\tilde{z}_t;\phi),
x_{t+1} = x_t-\tilde{x}_t\text{, }\tilde{x}_1=0,\\
\text{Minimize } \frac{1}{T}&\sum_{t=1}^{T}\mathcal{L}(x_1,\sum_{i=1}^{t}\tilde{x}_i).
\end{align}

\subsection{Deep Distributed Source Coding Framework}
Fig.~\ref{fig_0} and \ref{fig_1} illustrate our Distributed Recurrent Autoencoder for Scalable Image Compression (DRASIC). Similar to classical DSC framework, each data source is encoded separately and decoded jointly. In our network, each distributed encoder in Fig.~\ref{fig_0} has the exact same structure in Fig.~\ref{fig_1}. Traditionally, researchers have to design different kind of codes for specific data sources \cite{schonberg2004distributed}. We propose to use data-driven approach to handle complex scenarios where the distribution of data sources is unknown and their correlations can be arbitrary. Our proposal may also shed new light on sophisticated application scenarios such as videos where data sources and correlations are time dependent. 

In our neural network-based DSC, $M$ distributed encoders encode corresponding data sources $x^m$ that can be arbitrarily correlated. Each neural network-based encoder $f(x^m;\theta^m)$ has their own model parameters $\theta^m$. After binarizing bottleneck representations $z^m$, code sources $\tilde{z}^m$ are transmitted and concatenated batch-wisely. A single decoder $g(\tilde{z}^m;\phi)$ reconstructs images $\tilde{x}^m$ from all sources with the same model parameters $\phi$. In classical settings, the joint decoder has to process all compressed codes from each source jointly. In our data-driven setting, the joint training process optimizes the model such that the single decoder can decode from correlated sources. In this case, decoding codes from a particular data source does not depend on synchronization of codes from other sources, since the model has been optimized to adapt the correlations among all sources.
\begin{align}
z_t^m &= f(x_t^m;\theta^m),
\tilde{z}_t^m = \text{Binarize}(z_t^m),\\
\tilde{x}_t^{m} &= g(\tilde{z}_t^{m};\bm{\phi}),
x_{t+1}^{m} = x_t^{m}-\tilde{x}_t^{m}\text{, }\tilde{x}_1^{m}=0,
\end{align}
\begin{align}
\text{Minimize } \frac{1}{MT}&\sum_{t=1}^{T}\sum_{m=1}^{M}\mathcal{L}(x_1^m,\sum_{i=1}^{t}\tilde{x}_i^m).
\end{align}
Our result shows that the resulting distributed model can perform as well as encoding all data by one single encoder. However, if we encode and decode each data source separately, the performance becomes significantly worse, i.e. with $\tilde{x}_t^{m} = g(\tilde{z}_t^{m};\bm{\phi^m})$.

\section{Experiments}
\begin{figure}
\begin{subfigure}{0.5\textwidth}
    \centering
    \vspace{-0.2cm}
    \includegraphics[width=0.8\linewidth]{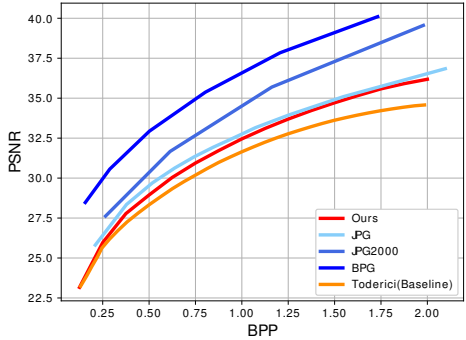}
    \vspace{-0.2cm}
    \caption{PSNR vs. BPP on Kodak dataset}
    \label{fig_3}
\end{subfigure}
\begin{subfigure}{0.5\textwidth}
    \centering
    \vspace{-0.2cm}
    \includegraphics[width=0.8\linewidth]{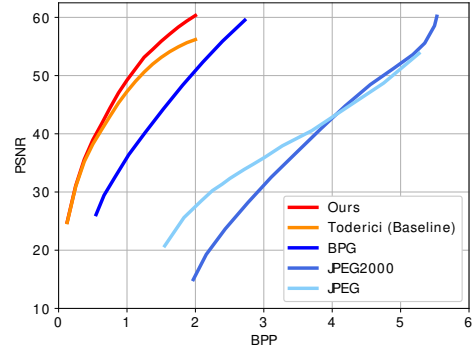}
    \vspace{-0.2cm}
    \caption{PSNR vs. BPP on MNIST dataset}
    \label{fig_4}
\end{subfigure}
\caption{Our symmetric recurrent autoencoder performs comparable to classical codecs and neural network-based codecs.}
\end{figure}

To show our model is capable of compressing natural images, we train our model on CIFAR10 dataset \cite{stojanovic2009underwater} and evaluate the rate-distortion curve on Kodak dataset \cite{Franzn2002KodakLT}. To show our model is capable of compressing grayscale images and demonstrate the feasibility of training encoders in a distributed manner, we train and evaluate our models with MNIST dataset \cite{lecun1998gradient}. We observe that many non-recurrent autoencoders outperform recurrent models on rate-distortion curves \cite{li2018learning,mentzer2018conditional}. We emphasize the distinction between the recurrent and non-recurrent autoencoders which do not have the scalability of reconstructing low quality images by using the subset of codes for high quality reconstruction. Our experiments aim to empirically demonstrate the feasibility of scalable distributed source coding in a data-driven setting. We use Adam optimizer \cite{kingma2014adam} with minibatch size of 100 for all experiments. We use learning rate $0.001$ for a total of 200 epochs and decay every $50$ epochs by a factor of $0.5$. Fig.~\ref{fig_3} shows that our symmetric recurrent autoencoder performs comparable to classical codecs and neural network-based codecs on compressing natural images, and performs significantly better on compressing handwritten grayscale images.

\begin{figure}
\begin{subfigure}{0.5\textwidth}
    \centering
    \vspace{-0.2cm}
    \includegraphics[width=0.8\linewidth]{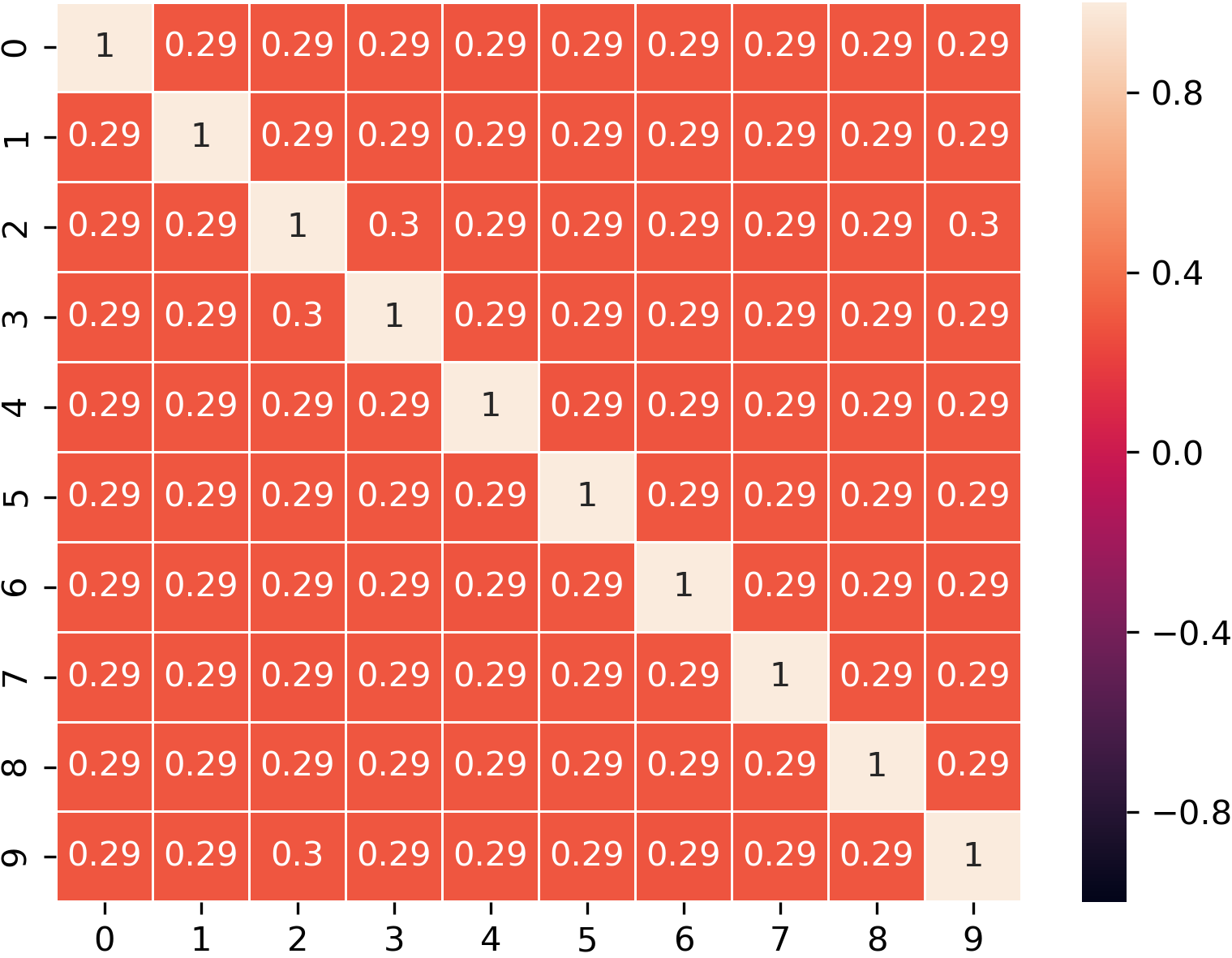}
    \vspace{-0.2cm}
    \caption{Split by random subsets.}
    \label{fig_5}
\end{subfigure}
\begin{subfigure}{0.5\textwidth}
    \centering
    \vspace{-0.2cm}
    \includegraphics[width=0.8\linewidth]{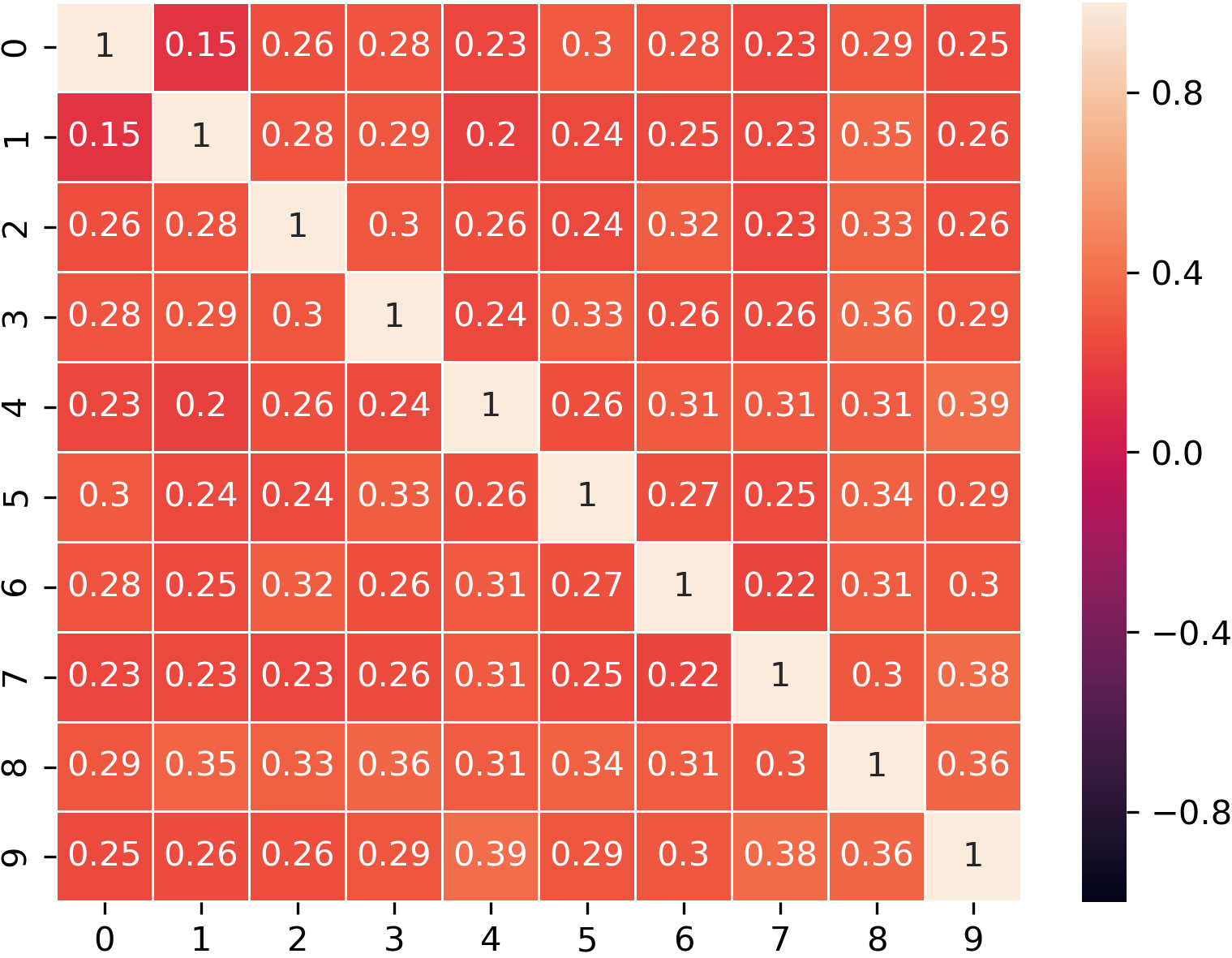}
    \vspace{-0.2cm}
    \caption{Split by class labels.}
    \label{fig_6}
\end{subfigure}
\caption{Pearson's correlation matrix among MNIST dataset.}
\end{figure}

To demonstrate the feasibility of compressing distributed data sources, we split our data into correlated subsets to emulate the case where encoders only have access to distributed correlated data sources. We conduct our experiments with $(2,4,8,10)$ number of distributed sources. For the MNIST dataset, the correlated data sources are from images separated by class labels. Each data source only contains the images of the same digit. First, we compare our result, labeled as \textit{Distributed}, to the case where all data are trained with one encoder and one decoder jointly, labeled as \textit{Joint}. The \textit{Joint} curve is approximated as the theoretical upper bound of performance. Second, we compare our result to the case where each data source is trained with a separate pair of encoder and decoder, labeled as \textit{Separate}. In Fig.~\ref{fig_5} and \ref{fig_6}, we illustrate the Pearson's correlation matrix among MNIST images split by random subsets and labels. It shows that the pixels of MNIST images are moderately correlated. Inspired by DSC, it is therefore possible to take advantage of their dependencies by training distributed encoders and a joint decoder. Our experimental studies in the following sections consist of three aspects. We first experiment $(2,4,8,10)$ number of distributed data sources with different correlations. We then show the robustness of our distributed framework in the absence of a number of distributed sources.

\begin{figure*}
\begin{center}
\includegraphics[width=0.8\linewidth]{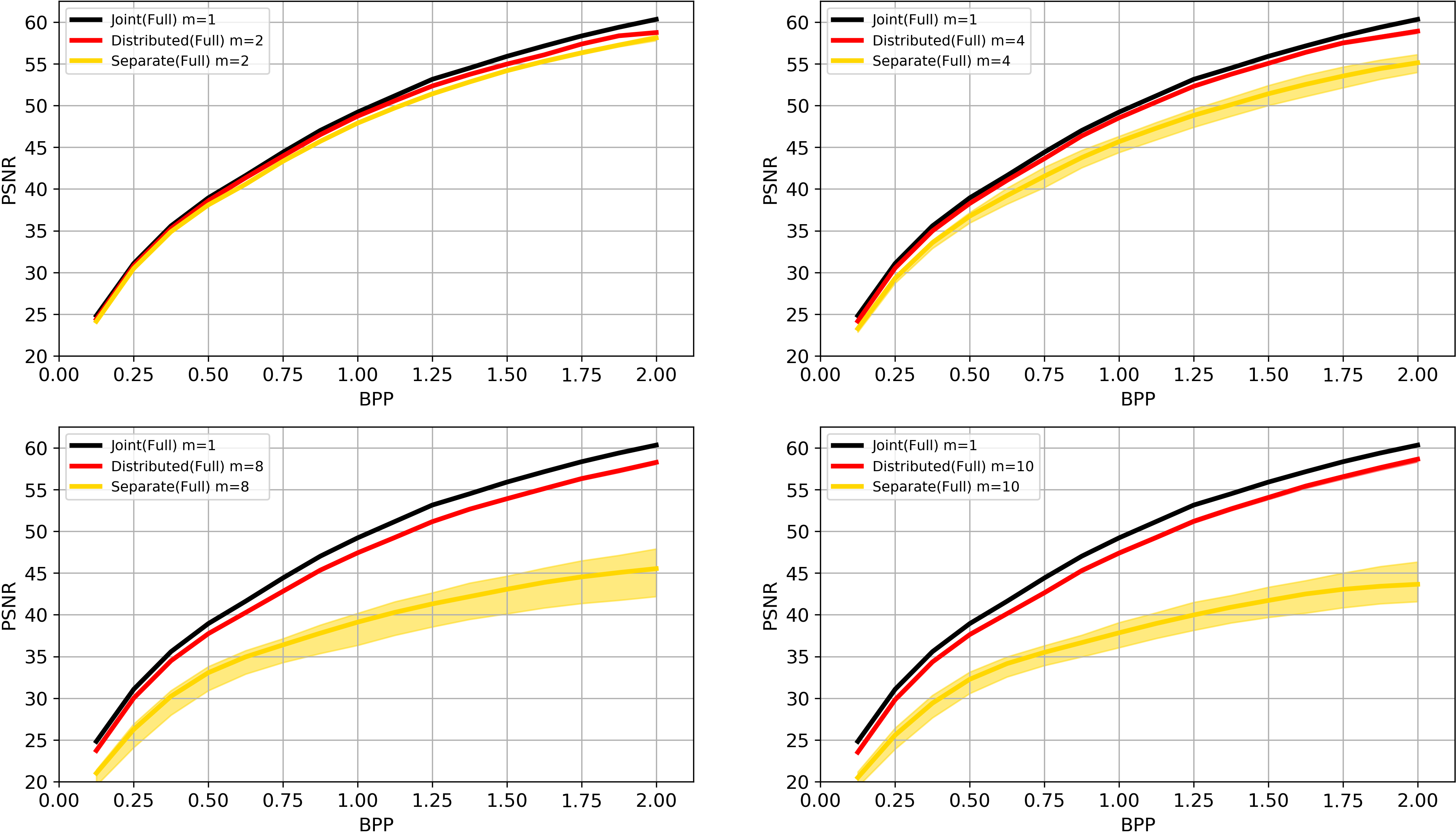}
\end{center}
	\vspace{-0.2cm}
   \caption{Rate-distortion curves for data sources distributed by random subsets with $T=16$ for all sources.}
   \vspace{-0.2cm}
\label{fig_7}
\end{figure*}

\begin{figure*}
\begin{center}
\includegraphics[width=0.8\linewidth]{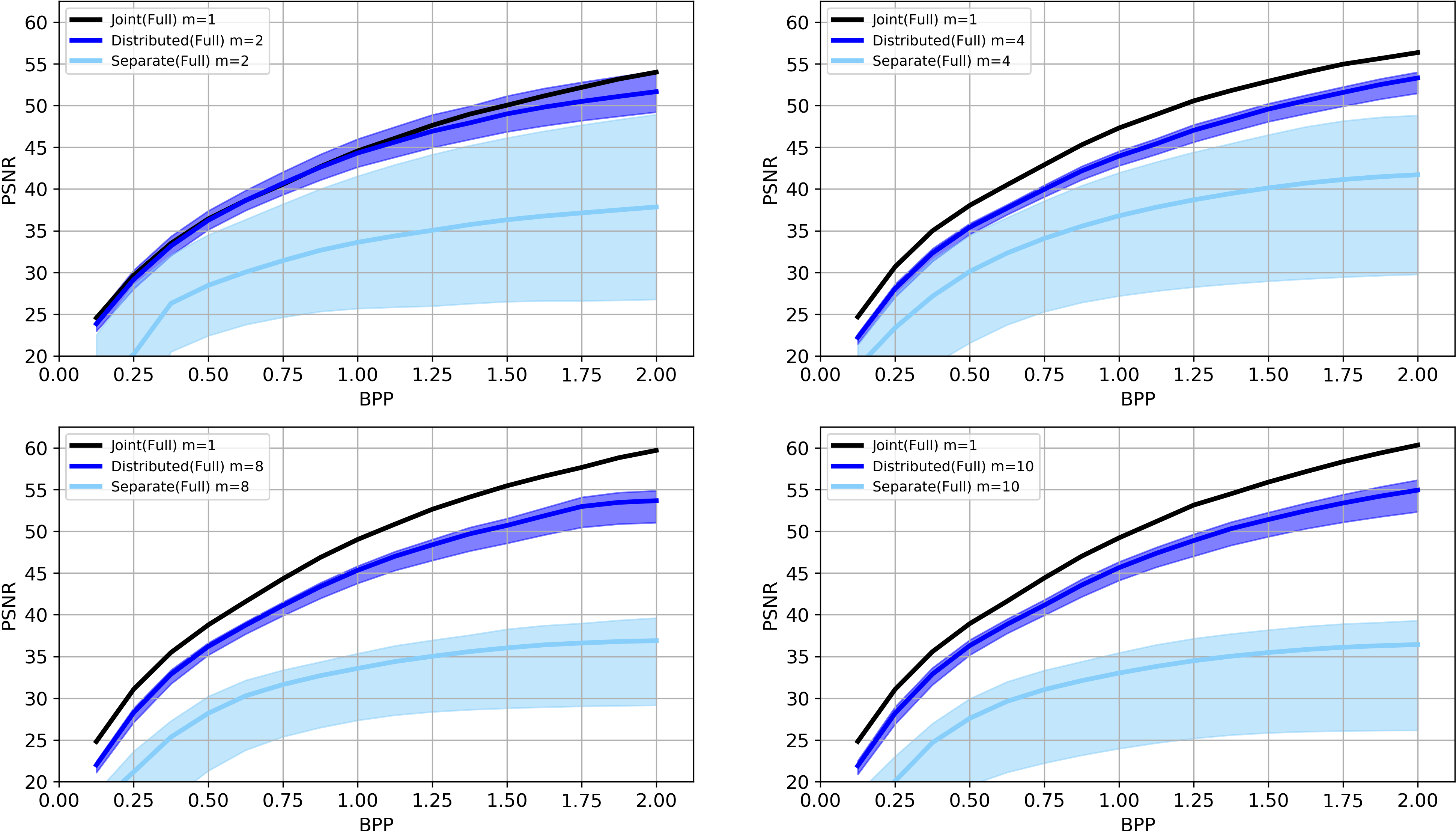}
\end{center}
	\vspace{-0.2cm}
   \caption{Rate-distortion curves for data sources distributed by class labels with $T=16$ for all sources.}
   \vspace{-0.2cm}
\label{fig_8}
\end{figure*}

To address the advantage of our DNN-based DSC framework, we experiment distributed sources with different correlations. The distributed encoders are labeled as $1,2, \ldots, m$. For example, when $m=2$, we only use first two subsets of images of digit $0$ and $1$. We show the result of data sources distributed by random subsets in Fig.~\ref{fig_7} and by class labels in Fig.~\ref{fig_8}. The curves of distributed encoders show that the performance of training distributed encoders and joint decoder can be very close to the theoretical limit. As the number of encoders grows, the performance decreases a little, but still dominantly outperforms training codecs for each data source separately. Results of images split by random subsets also outperform images split by class labels, it may relate to the constant correlation as shown in \ref{fig_5}. The results show that our Deep DSC framework can benefit from dependencies among an arbitrary number of data sources. 
Our data-driven DSC framework, unlike classical DSC code design, once deployed, does not require synchronization of data sources. In classical DSC code design, if syndrome bits $H(X|Y)$ are used and the data source $Y$ is accidentally blocked, we will not be able to decode the data source $X$. In our data-driven framework, even only one of the distributed encoders is functional, it can still benefit from its dependencies with other sources because their dependencies are already trained by the model parameters. All our experiments show that distributed encoders not only dominate separately trained codecs but also have narrower confidence bands. As the number of encoders increases, the confidence bands of separately trained codecs become wider because each separate codec can only access very limited amount of data and thus suffer from overfitting.

%------------------------------------------------------------------------
\section{Conclusion}
We introduced a data-driven Distributed Source Coding framework based on Distributed Recurrent Autoencoder for Scalable Image Compression (DRASIC). Compared with classical code designs, our method has the following advantages. First, instead of explicitly estimating the correlations among data sources before designing codes, the proposed data-driven approach can \textit{simultaneously learn the dependencies and compress}. Given enough training data, our method can handle an arbitrary number of sources with arbitrary correlations. Second, we showed the robustness of our framework. Unlike classical code designs which often require sophisticated data source synchronization, each distributed encoder of our model, once trained and deployed, can be used \textit{independently and asynchronously} of others. Each data source equipped with less data, fewer number of iterations, and smaller computational power can still approach the theoretical limit of compression obtained by pulling all the data. Last but not least, our recurrent model can reconstruct images efficiently even at low compression quality.

We point out two interesting directions of future work. First, the compression quality of the proposed architecture can be further improved by introducing spatially adaptive weights over different iterations, e.g. by using context models for adaptive arithmetic coding. Second, the network architecture can be further extended to handle time-dependent data sources.
%------------------------------------------------------------------------

\Section{References}
{
\bibliographystyle{IEEEbib}
\bibliography{References}
}

\end{document}